\documentclass{src/ieee-oj-its}

\usepackage{cite}
\usepackage{amsmath,amssymb,amsfonts}
\usepackage{algorithmic}
\usepackage{algorithm}
\usepackage{graphicx,color}
\usepackage{textcomp}
\usepackage{stfloats}
\usepackage{colortbl}
\usepackage[numbers]{natbib}

\def\BibTeX{{\rm B\kern-.05em{\sc i\kern-.025em b}\kern-.08em
    T\kern-.1667em\lower.7ex\hbox{E}\kern-.125emX}}
\AtBeginDocument{\definecolor{ojcolor}{cmyk}{0.93,0.59,0.15,0.02}}
\def\OJlogo{\vspace{-12pt}\includegraphics[height=24pt]{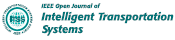}}

\begin{document}

\title{PAN: Pillars-Attention-Based \\ Network for 3D Object Detection}
\author{RUAN~BISPO\authorrefmark{1}~(Member,~IEEE),
        DANE~MITREV\authorrefmark{2},
        LETIZIA~MARIOTTI\authorrefmark{2},
        CLÉMENT~BOTTY\authorrefmark{2},
        DENVER~HUMPHREY\authorrefmark{2}~(Senior Member,~IEEE),\\
        ANTHONY~SCANLAN\authorrefmark{1},
        CIARÁN~EISING\authorrefmark{1}~(Senior Member,~IEEE)
}
        
\affil{Department of Electronic and Computer Engineering, Lero (the Research Ireland Centre for Software), and the Data Driven Computer Engineering (D$^2$iCE) Research Centre at the University of Limerick, Limerick, V94 T9PX Ireland.}
\affil{Provizio, Future Mobility Campus Ireland, Shannon Free Zone, V14WV82, Ireland.}

\corresp{CORRESPONDING AUTHOR: Ruan~Bispo (e-mail: bispodos.ruan@ul.ie).}
\authornote{This work was supported in part by Lero, (the Research Ireland Centre for Software) under Grant 13/RC/2094\_P2; and in part by Provizio.}
\markboth{PAN: Pillars-Attention-Based Network for 3D Object Detection}{Bispo et al.}

\begin{abstract}
Camera-radar fusion offers a robust and low-cost alternative to Camera-lidar fusion for the 3D object detection task in real-time under adverse weather and lighting conditions. However, currently, in the literature, it is possible to find few works focusing on this modality and, most importantly, developing new architectures to explore the advantages of the radar point cloud, such as accurate distance estimation and speed information.
Therefore, this work presents a novel and efficient 3D object detection algorithm using cameras and radars in the bird's-eye-view (BEV). Our algorithm exploits the advantages of radar before fusing the features into a detection head. A new backbone is introduced, which maps the radar pillar features into an embedded dimension. A self-attention mechanism allows the backbone to model the dependencies between the radar points. We are using a simplified convolutional layer to replace the FPN-based convolutional layers used in the PointPillars-based architectures with the main goal of reducing inference time. Our results show that with this modification, our approach achieves the new state-of-the-art in the 3D object detection problem, reaching 58.2 of the NDS metric for the use of ResNet-50, while also setting a new benchmark for inference time on the nuScenes dataset for the same category. 
\end{abstract}


\maketitle
\section{INTRODUCTION}\label{sec:introduction}
Through the current imperative for advancements in the domain of autonomous vehicles, various fields have evolved and gained significant prominence in recent years, with 3D object detection being a notable example \cite{qian20223d}. By leveraging camera images, lidar point clouds, or radar distance data as sensory inputs, 3D detection primarily aims to identify and locate objects such as pedestrians, cyclists, and other vehicles in the environment \cite{Wang2020}. This detection is conducted through the creation of three-dimensional bounding boxes that approximate the detected object with a simplified model of similar dimensions to the object in question.

Modeling objects using bounding boxes enables the differentiation of object types, such as pedestrians and vehicles, which exhibit distinct velocities and behaviors across various path segments and environmental conditions. By capturing these differences, bounding box-based modeling enhances the ability to anticipate interactions and mitigate potential risks in dynamic scenarios. As a result, the field of 3D object detection has experienced remarkable advancements in recent years, driven by innovative technologies \cite{mao20233dobjectdetectionautonomous}.

The task of 3D object detection is fundamentally dependent on the quality and characteristics of the sensing technologies used in the vehicle. The range of available sensors, such as cameras, lidar, and radar, has limitations due to noise, adverse environmental conditions, lighting variations, or even the intrinsic characteristics of each sensing device \cite{wang2020high}. Consequently, relying on a single sensor has proven to be a less viable solution, highlighting the necessity of integrating multiple sensors to leverage the advantages of one sensor in a manner that complements the capabilities of others \cite{fayyad2020deep}. 

Among the various challenges associated with sensing in 3D object detection, the limitations of sensors under adverse conditions stand out. In general, sensing systems are designed to operate under specific conditions, which are often highly restricted to controlled environments, excluding several important conditions. Numerous studies address the sensing problem under ideal conditions, such as infinite computational power, disregarding inference time, or even environments where sensors are not hindered by factors such as lighting, dust, fog, rain, or other environmental challenges \cite{pfeuffer2018optimal,li2022bevformer,nabati2021centerfusion}.

The limited availability of datasets encompassing adverse conditions has also contributed to the predominance of 3D detection studies focused on more controlled scenarios. Additionally, challenges such as processing data from multiple sensors in real-time and the need for a greater variety of sensors in datasets to support sensor fusion research have only begun to be addressed in recent years. Consequently, it has been observed that the perception field under adverse conditions remains relatively under-explored and requires further study to meet the growing demands generated by the development of autonomous vehicle projects \cite{Bijelic2020}.

This paper proposes a new architecture that uses camera-radar fusion as a detection modality for 3D object detection. The proposed approach aims to address the challenges of adverse weather conditions and a real-time approach by integrating complementary sensor data to enhance robustness and reliability. Furthermore, this study presents detailed experimental results evaluated against established reliability standards \cite{kim2023crn,li2023bevdepth,lin2024rcbevdet} and potential minimum requirements necessary for the adoption of this technology within the autonomous vehicle domain in real-world scenarios \cite{edquist2012effects}. 

\subsection{SUMMARY OF OUR MAIN CONTRIBUTIONS}

\begin{itemize}
    \item We are proposing PAN, a real-time camera-radar sensor fusion architecture, achieving around 30 FPS using the ResNet-50 as camera backbone.
    \item  We propose a pillars-attention-based features extractor backbone, which operates on the radar point cloud to extract features to be used in BEV tasks.
    \item We achieve state-of-the-art results using ResNet-18 and ResNet-50 on the 3D object detection benchmark nuScenes \cite{nuscenes2019} validation dataset even under subsets of adverse weather conditions.
    \item We show that PAN maintains robust performance under adverse weather conditions, such as rain and night scenarios, achieving state-of-the-art in both of these subsets.
\end{itemize}

\section{RELATED WORK}\label{sec:related_work}
\subsection{CAMERA-BASED 3D OBJECT DETECTION}

One of the best-known approaches to 3D object detection is to use only cameras. A camera dependency is due to our current paradigms of traffic signs and driving, which are based on visual and colored information \cite{contreras2024survey}. Among the camera-based methods, the BEVFormer \cite{yang2022bevformerv2adaptingmodern}, BEVDet \cite{huang2021bevdet}, and BEVDepth \cite{li2023bevdepth} are the most used so far because of their good cost-benefit between the metrics (e.g., 47.5 in the nuScenes Detection Score (NDS) \cite{nuscenes2019} for BEVDepth, using ResNet-50 as image backbone) and inference time. However, those methods still have low scores in terms of NDS, low scores under adverse weather conditions, and a low inference time. Currently, other methods are proposing solutions to address these problems by changing the structure of the camera network \cite{2305.14018,zou2024unim2aemultimodalmaskedautoencoders,liu2024raydenoisingdepthawarehard,huang2022bevpoolv2cuttingedgeimplementationbevdet,wang2023exploringobjectcentrictemporalmodeling}, but it is still a challenge in the literature.

Regarding techniques used to address adverse conditions, the work of \citeauthor{mehra2020reviewnet} stands out for tackling weather conditions involving foggy environments \cite{mehra2020reviewnet}. They propose ReViewNet, a method focused on camera data that addresses the issue through image dehazing to improve object detection. This approach is suitable for real-time use due to its fast processing nature, outperforming both CNNs and GAN-based architectures in terms of accuracy and processing speed. \citeauthor{pfeuffer2018optimal} \cite{pfeuffer2018optimal} employed a dataset transformation approach using the KITTI dataset \cite{geiger2012we}, a classic and widely used dataset in autonomous vehicle studies. In their work, modifications were incorporated to simulate adverse conditions focused on camera data, but some simulations were also used for the lidars. 

\subsection{POINT-BASED 3D OBJECT DETECTION}

The lidar sensor is currently the best sensor in terms of accuracy, 74.5 NDS, and there is a wide range of work in the literature using this sensor \cite{mei2025segtgeneralspatialexpansion, zhan2023realaugrealisticscenesynthesis, liu2024lionlineargrouprnn, mei2025segtgeneralspatialexpansion, zhang2024voxel}. However, the main disadvantage of this sensor is its extremely high cost \cite{cui2023redformer}, in addition to not being very robust against rain, lighting, and time of day variations \cite{nabati2021centerfusion}. One of the most used methods in this category is Point-Pillars \cite{lang2019pointpillars} due to its high inference time, metrics, and versatility as a standalone backbone to extract the shape-based features. This method uses a convolutional structure to extract the main features using a multi-scale approach.

Considering the radar-only approaches, few works are trying to improve this modality in the nuScenes benchmark \cite{ulrich2022improved, svenningsson2021radar}. Overall, despite radar being the best sensor in terms of robustness against adverse weather and lighting conditions, the small number of works is due to the current lack of good radar sensors available using sensors with better resolution and point density \cite{bijelic2020seeing}. The RadarDistill \cite{bang2025radardistillboostingradarbasedobject} is nowadays the best option in the nuScenes benchmark; it was the first architecture to achieve a metric of 43.7 in terms of NDS using the lidar data to improve the radar point density in the training step.

\subsection{SENSOR-FUSION FOR 3D OBJECT DETECTION}

In the context of sensor fusion, each sensor complements the information from the others to ensure the most accurate and reliable perception of the environment. Sensors such as cameras and lidars are highly susceptible to interference from adverse weather and lighting conditions, resulting in their data being partially or completely unusable in such scenarios, despite their high resolution \cite{qian20223d}. In contrast, radars exhibit greater immunity to these adverse conditions, even with a low resolution.

\citeauthor{Sheeny2021} presented the Radiate dataset, which was developed, containing radar, stereo camera, lidar, and GPS data collected simultaneously in various weather conditions \cite{Sheeny2021}. The study also compared the performance of models trained with radar data, highlighting the minimal interference of weather on radar performance. Furthermore, the work of \citeauthor{bijelic2020seeing} is one of the few works focused on sensor fusion in adverse weather conditions, proposing a method and dataset to address this problem, including fog, rain, and snow \cite{bijelic2020seeing}.

Focusing on using cameras and radar exclusively, some works, such as REDFormer \cite{cui2023redformer}, propose camera-radar sensor fusion to improve general detection metrics, including performance under adverse conditions. This is achieved by incorporating radar feature extraction networks and a multi-task learning approach to identify adverse conditions, though at the expense of increased processing time. Other studies aim to address the alignment problem between radar and camera data, as seen in MVFusion \cite{wu2023mvfusion}, but similarly incur higher processing demands and increased data volume.

Additionally, some studies have attempted to integrate radar data with lidar data during the training phase or by generating depth maps. These methods currently represent the state-of-the-art in terms of metrics on the official nuScenes benchmark, achieving detection performance comparable to that obtained with lidars, with scores exceeding 50.0 in the nuScenes Detection Score (NDS) \cite{lei2023hvdetfusion, kim2023crn, lei2023hvdetfusionsimplerobustcameraradar}. However, this approach has the drawback of not exploring the radar features so deeply.

Recent advances in sensor fusion techniques using a camera-radar approach \cite{lin2024rcbevdet} aim to leverage radar features more effectively to complement the camera's gaps through a new radar backbone proposal. However, despite the unique characteristics of cameras (e.g., inaccurate BEV transformation) and radars (e.g., sparsity and ambiguity), combining the advantages of both is nontrivial \cite{kim2023crn}. Furthermore, this method relies on a complex architecture and does not primarily focus on performance under adverse weather conditions. 

\section{ARCHITECTURE}\label{sec:architecture}
\begin{figure*}[!t]
\centering
\includegraphics[width=7in]{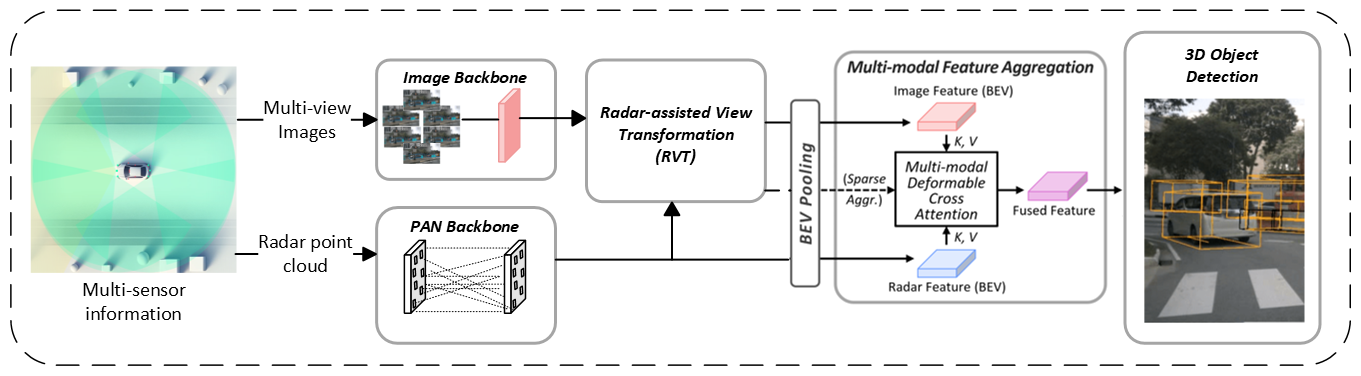}
\caption{Overall architecture proposed. Firstly, the radar features are extracted using the PAN backbone and in parallel, camera features are extracted using an image backbone. Secondly, we transform these features into a BEV using the radar features through the Radar-assisted view transformation. Finally, the Multi-modal Feature Aggregation works by merging radar and camera BEV features to feed the 3D detection head.}
\label{fig_overview}
\end{figure*}

In this paper, we propose a camera-radar architecture for 3D object detection, leveraging a middle fusion technique that combines the camera and radar-extracted features using a transformer-based fusion in the bird’s-eye-view. In previous works presented in the literature, we observed that we could further exploit the radar sensor features using attention mechanisms and that their use has become more frequent in robotics and other perception tasks to improve the accuracy and reduce the inference time \cite{de2022attention}. Furthermore, we are proposing a camera-radar fusion because of its main advantage of robustness under adverse weather conditions, time-of-day variations, and its low cost when compared to lidar-based sensor fusion techniques.

As shown in Fig. \ref{fig_overview}, the Pillars-Attention-Based Network (PAN) for 3D Object Detection consists of four main stages. Firstly, the camera and radar information are used to feed a modality-specific encoder to extract their features; ResNet was used for the image data, and the PAN backbone was used for the radar data. Secondly, using these features, we apply the radar-assisted view transformation (RVT) to convert the image features from a camera perspective into a bird's-eye view (BEV). In this component, we use a convolutional head to convert the radar features into a binary occupancy map. In the 3rd stage, the radar and camera BEV features are fused using the multi-modal Feature Aggregation module, where a cross-attention mechanism is applied to unify the BEV features. Lastly, the features are used to feed the 3D detection head.

In the same way that it is common in the literature to extract image features using pre-trained ResNet architectures \cite{he2016deep}, in recent years, the main architecture used for point-based data, such as lidar, has been Pointpillars \cite{lang2019pointpillars}. This architecture is highly dependent on the number of points since it primarily uses 2D convolution to process the point cloud shape. The computational complexity of convolution scales with \begin{math}O(n)\end{math}, where in \begin{math}n\end{math} is made up of the number of filters, filter size, and mainly the dimension of the data.
In the radar case where the points are very sparse, many convolutions are performed in empty spaces \cite{musiat2024radarpillarsefficientobjectdetection}. To overcome this problem, we propose a Pillars-attention-based network to effectively extract the radar features and fuse them with camera features, excluding the empty spaces and better using the radar features, thereby improving the performance in terms of metrics and reducing the computational cost.

\subsection{PAN BACKBONE}

\begin{figure}[!htbp]
\centering
\includegraphics[width=3.4in]{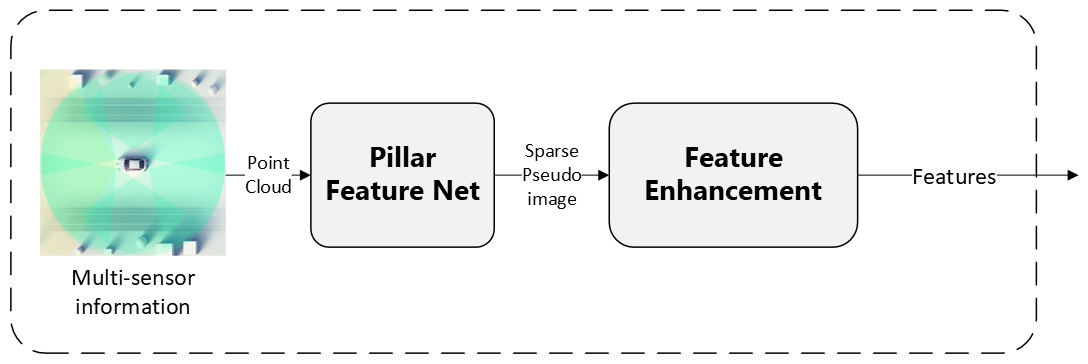}
\vspace{-3mm}
\caption{Summary for the PAN backbone showing how this module can be used for this architecture or as a decoupled module for other perception tasks.}
\label{fig_pan_backbone}
\end{figure}

The Pillars-Attention-based Backbone can be used as a 3D object detection network, as we are proposing in the paper, or it can be used as a feature extractor module to address other perception problems (similar to  Pointpillars \cite{lang2019pointpillars}), which can be used as 3D detector, or as a backbone to deal with other tasks as a decoupled block, where each task is evaluated by different works producing different levels of results \cite{liu2022bevfusion,harley2023simple,chen2022proposal}. A summary of this application is shown in Fig. \ref{fig_pan_backbone}, where the input is a multi-sensor point cloud using five radars to cover 360 degrees of view, and the output is the BEV feature components.

Initially, the radar point cloud using all radars available and previous data frames in the camera view, \begin{math} pc \in \mathbb{R}^{H\times W \times C_{r}} \end{math}, respectively depending on the radar distance, camera width, and radar features, are fed into a pillar feature net block \cite{qi2017pointnet++} to convert the point cloud into a sparse pseudo-image with the first high-dimensional features, \begin{math} x \in \mathbb{R}^{H \times W \times C} \end{math}, extracted. Note that for nuScenes' radar data, we are not performing the operations in the camera height dimension since the available radar does not provide reliable elevation data.  After that, these pseudo-images are fed into a feature enhancement block that removes the sparsity, refines the extracted features using a self-attention mechanism, followed by the convolutional layers, and recomposes the 2D grid shape.

\begin{figure}[!htbp]
\centering
\includegraphics[width=3.2in]{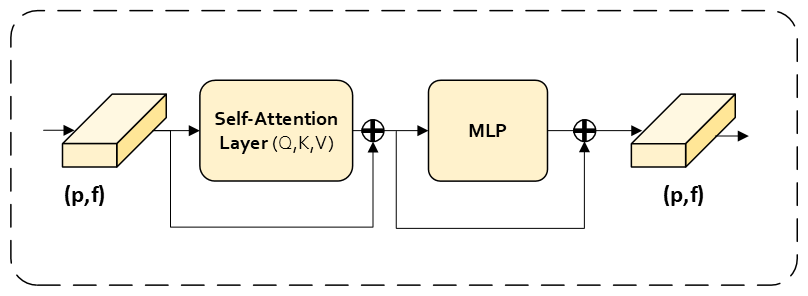}
\caption{Self-attention branch diagram. The self-attention component can benefit from radar features such as sparse but accurate speed and distance to improve metric results while using low computational resources due to the nature of the data.}
\label{self_att_zoom}
\end{figure}

\begin{figure*}[!htbp]
\centering
\includegraphics[width=6.8in]{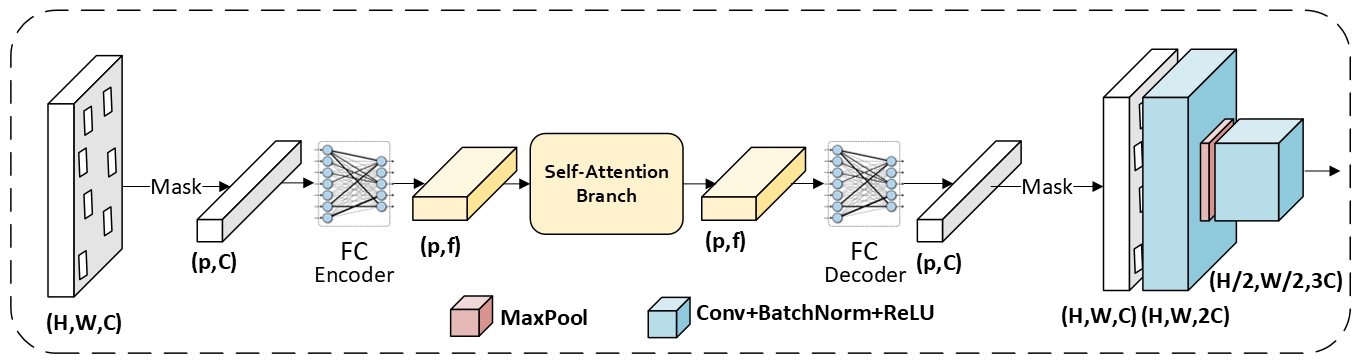}
\caption{Detailed Feature Enhancement Block diagram, where the input is the sparse pseudo-image \begin{math} (H,W,C) \end{math}, and the output is the enhanced features. Given a sparse pseudo-image with radar features, a mask is applied to remove all empty pillars. Encoding reduces the dimensionality of the features. A self-attention branch is sequentially applied to capture the most highly correlated features. After a fully connected decoder, a simplified convolutional block is applied to refine the close feature correlations and adjust the pseudo-image shape.}
\label{fig_att_block}
\end{figure*}

The feature enhancement block is composed of a first step where a sparsity mask is applied to each input frame to remove the empty pillars. In this step, the \begin{math}H\end{math}, and \begin{math}W\end{math} dimensions are combined into a single dimension \begin{math}p\end{math} for normalization and the next features refinement steps. Using a simple, fully connected layer applied to each non-empty pillar, \begin{math} p \end{math}, all the features are encoded to the embedded space with dimension \begin{math} f \end{math}. Sequentially, the features are refined using the self-attention branch and decoded using the same structure as the decoder. Finally, the convolutional component is applied to adjust and refine the features. The self-attention component is exemplified using Fig. \ref{self_att_zoom}. A complete diagram for the feature enhancement block is shown in Fig. \ref{fig_att_block}

The key component for the feature enhancement block is a self-attention branch to model dependencies between different parts of the radar data, where the encoded features \begin{math} f \end{math} are used to calculate the \begin{math} Q, K \end{math}, and \begin{math} V \end{math} values through three linear layers. Afterwards, the attention scores and values are calculated, enabling the model to identify important relationships among sequence elements. We are using the self-attention approach proposed by \cite{vaswani2017attention}, exemplified in \eqref{eq_selfatt}.
\begin{equation}
\label{eq_selfatt}
x_{(p,f)}(Q,K,V) = softmax(\frac{QK^T}{\sqrt{d_k}})V.
\end{equation}
where \begin{math} x \in \mathbb{R}^{p \times f} \end{math} are the enhanced features,  \begin{math} Q, K \end{math}, and \begin{math} V \end{math} are the query, key, and value variables respectively of the transformers model, and \begin{math} \sqrt{d_k} \end{math} is the attention normalization factor, used to prevent the similarity values from becoming excessively large, in which \begin{math} d_k \end{math}, is the dimensionality of the keys \begin{math} K \end{math}. Furthermore, we are applying a dropout layer \cite{JMLR:v15:srivastava14a} for normalization and reducing the overfitting before the softmax function \cite{bridle1990probabilistic}.

The attention results \begin{math} x_{(p, f)} \end{math} are then added to the encoded features to feed a two-layer MLP (Multi-Layer-Perceptron) module, which is composed of a linear layer, with a layer normalization and a GeLU as activation function, followed by a last linear layer, both using the same dimension of the features \begin{math} f \end{math}. The output from the MLP module and the residual component are added to feed the fully connected decoder layer. Similar to the encoder layer, the enhanced features are transformed back into a sparse image using the same sparsity mask to feed the convolutional components of the feature enhancement block. A two-stage convolutional block further processes the resulting sparse image with the enhanced features to improve the local features and adjust the pseudo-image size. 

The first convolution step is formed using a 2D Batch Normalization \cite{ioffe2015batch} and ReLU as an activation function. A Maxpooling layer \cite{lecun1989handwritten} is applied to reduce the pseudo-image size. A second convolutional layer is used to capture some spatial features better, and the channel depth of the feature maps is increased to retain informational content. As the self-attention step was used to extract the feature dependencies between far radar points, it is necessary to use only a few convolutional layers to adjust the final features. A pseudo-algorithm is shown at Algorithm \ref{alg:alg1} for reproducibility purposes.

\begin{algorithm}[ht]
\caption{PAN Backbone.}\label{alg:alg1}
\begin{algorithmic}
\STATE 
\STATE {\textsc{BACKBONE}}$(\mathbf{Point Cloud})$
\STATE \hspace{0.5cm}$ \textbf{convert to sparse pseudo image}$
\STATE \hspace{1.0cm}$ \textbf{x} (bs, h, w, c) \gets \textbf{pc} (bs, h, w, c_{r})$
\STATE \hspace{0.5cm}$ \textbf{feature refinement}$
\STATE \hspace{1.0cm}$ \textbf{x} (pc, c) \gets \textbf{x} (bs, h, w, c)[mask]$
\STATE \hspace{1.0cm}$ \textbf{x} (pc, f) \gets \text{Linear}(\textbf{x} (pc, c))$
\STATE \hspace{1.0cm}$ \textbf{x} (pc, f) \gets \text{Self-Attention-branch}(\textbf{x} (pc, f))$
\STATE \hspace{1.0cm}$ \textbf{x} (pc, c) \gets \text{Linear}(\textbf{x} (pc, f))$
\STATE \hspace{1.0cm}$ \textbf{x} (bs, h, w, c)[mask] \gets \textbf{x} (pc, c)$
\STATE \hspace{1.0cm}$ \textbf{x} (bs, h, w, 3c) \gets \text{Conv}(\textbf{x} (bs, h, w, c))$
\end{algorithmic}
\label{alg1}
\end{algorithm}

\subsection{RADAR-ASSISTED VIEW TRANSFORMATION}

For the camera-view to BEV conversion, we use the radar-assisted View Transformation \cite{kim2023crn}, combining radar and camera data to improve the transformation performance. It transforms the camera features from the camera view, adding a depth estimation using the radar occupancy and the depth distribution. This allows the accurate depth information from the radar to be fused with the visual richness of the camera images to improve the depth estimation. After this transformation, the BEV Pooling transformation is applied to the resultant camera and radar features through a CUDA-enabled Voxel Pooling \cite{li2023bevstereo}, unifying the features in the BEV space. In this case, as we have more precise features extracted and occupancy estimation using the PAN backbone, we can add more precise information to this transformation, increasing the final performance of the transformation.

\subsection{MULTI-MODAL FEATURE AGGREGATION}

During the Multi-modal Feature Aggregation (MFA), the camera and radar BEV features are fused using the Multi-modal Deformable cross-attention (MDCA) \cite{kim2023crn}, employing a cross-attention mechanism to merge the regions where camera and radar data are most correlated, to remove calibration and data misalignment issues. The equation for this mechanism can be expressed at \eqref{eq:cross_att}.
\begin{multline}
\label{eq:cross_att}
\mathrm{MDCA} = \\
\sum_{h=1}^{H} W_{h} \left[
    \sum_{m=1}^{M} \sum_{k=1}^{K}
    A_{hmqk} \cdot W'_{hm} \,
    x_{m} \big( \phi_{m}(p_{q} + \Delta p_{hmqk}) \big)
\right]
\end{multline}

where \(H\) is the number of attention heads, $M$ denotes the number of modalities and \(K\) is the number of sampling points per head. $W_{h}$ and $W'_{hm}$ are the output and input projection matrices, respectively, for head $h$ and modality $m$. $A_{hmqk}$ are the attention weights and $\phi_{m}$ scales the normalized coordinates $p_{q}$ for each query element in case two modalities have different shapes. The term $\Delta p_{hmqk}$ is the offset applied to the reference position.

\subsection{3D OBJECT DETECTION HEAD}

For the 3D object detection head, we employ the CenterPoint head \cite{yin2021center}, a robust and widely used anchor-free detection approach. This method first predicts a center heatmap using a multi-group head \cite{zhu2019class}, effectively locating object centers without relying on predefined anchors, which improves efficiency and adaptability. During the training phase, sparse aggregation is utilized to enhance performance by filtering out points that fall outside the 3D bounding boxes when generating the ground truth depth map using the LiDAR information. This ensures that only relevant points contribute to the model's learning process, reducing noise and improving accuracy. For consistency and optimal performance, we adopt the same configuration used in CRN \cite{kim2023crn}. 

\section{EXPERIMENTS}\label{sec:experiments}
In this section, we initially present our methodology, explaining the dataset we used and its main metrics of validation, as well as presenting our experimental results; sequentially, we show an analysis of some safety aspects in our work, guiding future works and suggesting some minimum requirements for a real-world implementation; we also present an ablation study subsection to show how the most important components of our work influence the final result; and finally, we present a qualitative results subsection to visually show how our work is contributing to improving 3D object detection results in the autonomous car domain.

\subsection{DATASET AND METRICS}

For this study, the nuScenes dataset \cite{nuscenes2019} was utilized due to its variety of environments, including nighttime, rainy, and daytime conditions, as well as its availability of both camera and radar data. Moreover, it offers a larger data volume and a good variety of scenarios compared to other current datasets. The nuScenes dataset includes 1x lidar, 6x cameras, and 5x radars with 18 channels, operating at frequencies of 20Hz, 12Hz, and 13Hz, respectively. All these sensors were positioned to capture information, covering a 360-degree field of view.

The nuScenes dataset employs the nuScenes Detection Score (NDS) as its primary evaluation metric for validating model performance. This metric combines the mean average precision (mAP) with a set of true positive metrics (mTP) that measure specific aspects of detection performance, including Average Translation Error (ATE), Average Scale Error (ASE), Average Orientation Error (AOE), Average Velocity Error (AVE), and Average Attribute Error (AAE). The calculation of the NDS is illustrated in \eqref{eq:nds}.

\begin{equation}
\label{eq:nds}
NDS = 0.5 \times mAP + 0.1 \times \sum_{mTP} (1 - min(1, mTP)).
\end{equation}

The entire nuScenes training dataset was used, and validation was performed by partitioning the validation set using its description in the metadata for rainy and nighttime data, in order to enable the study under adverse weather conditions and low visibility cases. The dataset distribution follows the percentage of 12.03\% and 19.84\% for night and rain scenarios, considering the train set, and 10.00\% and 18.08\% of the total validation set. The standard train and validation sets are divided in a proportion of around 80/20.

Finally, the image size chosen for this work was \begin{math} (256\times704) \end{math}, equivalent to approximately 13\% of the total image size available in pixels, on the nuScenes dataset \begin{math} (1600\times900) \end{math}. This adjustment was based on previous works, such as \cite{li2023bevdepth, liu2022bevfusion, huang2021bevdet, kim2023crn, lin2024rcbevdet} that reduced the image size to \begin{math} (256\times704) \end{math}, to reduce the computational cost and offer a massive increase in the inference time of around 9.2 times when compared to the full image size, without a major performance drop.

\subsection{IMPLEMENTATION DETAILS}

The PyTorch library, MMDetection3D toolbox \cite{mmdet3d2020}, and Python language were used to train and validate the model. The model was trained using a single NVIDIA GeForce RTX 4070-S GPU, 32GB of RAM, and Intel® Core™ i9-14900K Desktop Processor. We explored predefined hyperparameter ranges, tested combinations empirically, and selected the best-performing configuration, with key parameter tuning detailed as follows. We are training for 48 epochs, using the AdamW optimizer in an end-to-end manner, and the hyperparameters used during training are as follows: a baseline learning rate of \begin{math} 2e-4 \end{math}, reducing the learning rate by a factor of 0.1 in the 5th last and last epochs. Additionally, we are training without CBGS \cite{zhu2019class} and employing the previous six radar sweeps for the multi-sweep approach.

\subsection{EXPERIMENTAL RESULTS}

To evaluate the performance of the proposed architecture in 3D object detection, its validation was conducted using the nuScenes validation dataset (with the validation splits as discussed in Section~\ref{sec:experiments}.A), and the results were compared with baseline models. The analysis is focused on the NDS metrics, which are the primary evaluation metric for the nuScenes dataset, and secondly on the FPS, which indicates the inference time of the models. Table \ref{tab:results} presents our main results using the validation set, highlighting in gray the performance of our models with two different image backbones. Our best result was using the ResNet-50, reaching 58.2 of NDS (about 4\% improvement) and 29.2 FPS (about 43\% improvement) over our main base model.

\begin{table*}[ht]
    \centering
    \renewcommand{\arraystretch}{1.2}
    \caption{Comparison of different 3D detection methods on the validation set. Highlighted values in bold represent the best performance in the respective category, and L, C, and R represent lidar, camera, and radar, respectively. $\ast$: results from MMDetection3D [4]. †: trained with CBGS. We are reporting the results found in the paper for CRN.}
    \label{tab:results}
    \setlength{\tabcolsep}{6.4pt}
    \begin{tabular}{l|c|c|c||c||c|c|c|c|c|c|c}
        \hline
        Method & Input & Backbone & Image Size & \textbf{NDS}$\uparrow$ & mAP$\uparrow$ & mATE$\downarrow$ & mASE$\downarrow$ & mAOE$\downarrow$ & mAVE$\downarrow$ & mAAE$\downarrow$ & FPS$\uparrow$ \\
        \hline
        CenterPoint-P$^\ast$ \cite{yin2021center} & L & Pillars & - & 59.8 & 49.4 & 0.320 & 0.262 & 0.377 & 0.334 & 0.198 & - \\
        CenterPoint-V$^\ast$ \cite{yin2021center} & L & Voxel & - & 65.3 & 56.9 & 0.285 & 0.253 & 0.323 & 0.272 & 0.180 & - \\
        \hline
        BEVDet$^\dagger$ \cite{huang2021bevdet} & C & R50 & 256 $\times$ 704 & 39.2 & 31.2 & 0.691 & 0.272 & 0.523 & 0.909 & 0.247 & 15.6 \\
        CenterFusion$^\dagger$ \cite{nabati2021centerfusion} & C+R & DLA34 & 448 $\times$ 800 & 45.3 & 33.2 & 0.649 & 0.263 & 0.535 & 0.450 & 0.142 & - \\
        BEVDepth$^\dagger$ \cite{li2023bevdepth} & C & R50 & 256 $\times$ 704 & 47.5 & 35.1 & 0.639 & 0.267 & 0.479 & 0.428 & 0.198 & 11.6 \\
        RCBEV4d$^\dagger$ \cite{zhou2023bridging} & C+R & Swin-T & 256 $\times$ 704 & 49.7 & 38.1 & 0.526 & 0.272 & 0.445 & 0.465 & 0.185 & - \\
        CRAFT$^\dagger$ \cite{kim2023craft} & C+R & R50 & 448 $\times$ 800 & 51.7 & 41.1 & 0.494 & 0.276 & 0.454 & 0.486 & 0.176 & 4.1 \\
        \hline
        CRN \cite{kim2023crn} & C+R & R18 & 256 $\times$ 704 & 54.3 & \textbf{44.8} & 0.518  & 0.283 & 0.552 & 0.279 & \textbf{0.180}  & 27.9 \\
        RCBEVDet \cite{lin2024rcbevdet} & C+R & R18 & 256 $\times$ 704 & 54.8 & 42.9 & \textbf{0.502} & 0.291 & \textbf{0.432} & \textbf{0.210} & 0.178  & 28.3 \\
        \rowcolor[gray]{0.9}
        PAN (ours) & C+R & R18 & 256 $\times$ 704 & \textbf{55.3} & 44.4 & 0.506 & \textbf{0.281} & 0.452 & 0.262 & 0.186 & \textbf{35.6} \\
        \hline
        SOLOFusion$^\dagger$ \cite{park2022time} & C+R & R50 & 256 $\times$ 704 & 53.4 & 42.7 & 0.567 & \textbf{0.274} & 0.411 & 0.252 & 0.188 & 11.4 \\
        CRN \cite{kim2023crn} & C+R & R50 & 256 $\times$ 704 & 56.0 & \textbf{49.0} & 0.487 & 0.277 & 0.542 & 0.344 & 0.197 & 20.4 \\
        RCBEVDet \cite{lin2024rcbevdet} & C+R & R50 & 256 $\times$ 704 & 56.8 & 45.3 & \textbf{0.486} & 0.285 & \textbf{0.404} & \textbf{0.220} & 0.192 & 21.3 \\
        \rowcolor[gray]{0.9}
        PAN (ours) & C+R & R50 & 256 $\times$ 704 & \textbf{58.2} & 48.1 & 0.488 & 0.279 & \textbf{0.404} & 0.232 & \textbf{0.181} & \textbf{29.2} \\
        \hline
    \end{tabular}
\end{table*}

Compared to the baseline CRN, an improvement of approximately +35\% in total execution time was achieved, based on the average across both camera backbones. Reducing inference time is an essential criterion for future implementations of camera-radar systems in autonomous vehicles. Table \ref{tab:results} illustrates the performance of the main baseline considering only the camera, BEVDepth \cite{li2023bevdepth}, and the camera-radar fusion, CRN, presented in the original article with similar hardware, as well as the results for the proposed variation of the PAN architecture, considering both camera backbones.

The second evaluation approach used in our work was to separate the validation dataset using the rain and night categories. It is possible to observe our model reaching the state-of-the-art in each one of those categories, considering the main metric NDS in Table \ref{tab:weather_conditions}. This improvement may show that to improve performance under adverse weather conditions, where the camera does not work as well as the radars, a possible path may be to improve the use of radar information or technology. The results presented for the other architectures were the same as reported by the authors, except in the CRN model, where the reported results were reported using a different image size.

\begin{table}[ht]
    \centering
    \renewcommand{\arraystretch}{1.2}
    \caption{Performance comparison under different weather conditions. The values in parentheses represent the improvement over the CRN baseline. We are using the results reported by the authors except for CRN, where the results using the image size of 256x704 were not available.}
    \label{tab:weather_conditions}
    \setlength{\tabcolsep}{3.2pt}
    \begin{tabular}{l|c|cc|cc}
        \hline
        Method & Input & \multicolumn{2}{c|}{Rainy} & \multicolumn{2}{c}{Night} \\
        & & NDS$\uparrow$ & mAP$\uparrow$ & NDS$\uparrow$ & mAP$\uparrow$ \\
        \hline
        BEVDepth \cite{li2023bevdepth} & C & - & 30.0 & - & 16.8 \\
        BEVFormer \cite{li2022bevformer} & C & 38.8 & 35.2 & 19.1 & 18.2 \\
        RCBEV \cite{zhou2023bridging} & C+R & 50.0 & 38.5 & 23.7 & 15.5 \\
        REDFormer \cite{cui2023redformer} & C+R & 50.9 & 40.4 & 28.1 & 20.3 \\
        \hline
        CRN-R18 \cite{kim2023crn} & C+R & 52.9 & 43.3  & 31.2 & \textbf{24.2} \\
        \rowcolor[gray]{0.9}
        PAN-R18 (ours) & C+R & \textbf{57.6 (+8.6\%)} & \textbf{47.6} & \textbf{33.7 (+7.4\%)} & 23.8 \\
        \hline
        CRN-R50 \cite{kim2023crn} & C+R & 56.9 & 48.1 & 35.4 & \textbf{27.8} \\
        \rowcolor[gray]{0.9}
        PAN-R50 (ours) & C+R & \textbf{59.8 (+5.1\%)} & \textbf{49.7} & \textbf{36.1 (+2.0\%)} & 27.0 \\
        \hline
    \end{tabular}
\end{table}

We further compare our model using all ten classes, detailing the metrics that calculate NDS and metrics where our architecture showed more expressive variations, such as AOE and AAE, in table \ref{tab:complete_results}. The other attributes that are more dependent on the shape of the object, such as translation and scale, are expected to be a challenge to improve their estimation using radar-based improvements because of the low density of points. We believe this improvement can be attributed to the capability to gather better global information using transformers and radar features. This way, using the velocity vectors, we can better estimate the orientation (AOE) and the object attribute (AAE) representing some properties, such as an object moving or standing. The traffic cone and barriers do not have these metrics.

\begin{table*}[ht]
    \centering
    \renewcommand{\arraystretch}{1.2}
    \caption{Comparison of different 3D detection methods in the validation set, considering different metrics per class. The R50 version represents the version using ResNet-50 as the image backbone, and the metrics AOE and AAE are not computed for traffic cones and barriers as standard. The presented results for the other architectures were reported utilizing the values shown by the respective authors. Values in bold represent the best performance and those in italics the second best.}
    \label{tab:complete_results}
    \setlength{\tabcolsep}{6.2pt}
    \begin{tabular}{l|c|cccccccccc|c}
        \hline
        Method       & Metric & Car & Truck & Bus & Trailer & C. V. & Pedestrian & Motorcycle & Bicycle & Traffic Cone & Barrier & mean \\
        \hline
        CenterNet \cite{zhou2019objects}    & AP$\uparrow$             & 48.4         & 23.1           & 34.1         & 13.1             & 3.5          & 37.7          & 24.9          & 23.4             & 55.0         & 45.6            & 30.6         \\
        CenterFusion \cite{nabati2021centerfusion} & AP$\uparrow$           & 52.4   & 26.5     & 36.2   & 15.4       & 5.5    & 38.9   & 30.5    & 22.9       & 56.3   & 47.0      & 33.2   \\
        CRAFT-I \cite{kim2023craft}      & AP$\uparrow$             & 52.4         & 25.7           & 30.0         & 15.8             & 5.4          & 39.3          & 28.6          & 29.8             & 57.5         & 51.1            & 33.2         \\
        CRAFT \cite{li2023bevdepth}       & AP$\uparrow$           & 69.6  & 37.6    & 47.3  & 20.4      & 10.7   & 46.2    & 39.5   & 31.0       & 57.1   & 51.1      & 41.1   \\
        BEVDepth \cite{li2023bevdepth}     & AP$\uparrow$             & 55.3         & 25.2           & 37.8         & 16.3             & 7.6          & 36.1          & 31.9          & 28.6             & 53.6         & 55.9            & 34.8         \\
        CRN-R50      \cite{kim2023crn}             & AP$\uparrow$           & \emph{73.6}  & \textbf{44.5}   & \textbf{55.6}  & \emph{22.0}       & \textbf{15.4}   & \emph{50.2}  & \textbf{54.7}  & \textbf{48.9}     & \textbf{61.4} & \textbf{63.8}  & \textbf{49.0}  \\ 
        \rowcolor[gray]{0.9}
        PAN-R50 (ours) & AP$\uparrow$ & \textbf{73.9} & \emph{42.2} & \emph{54.8} & \textbf{25.1} & \emph{13.4} & \textbf{52.4} & \emph{52.7} & \emph{44.6} & \emph{59.1} & \emph{62.7} & \emph{48.1} \\
        \hline
        BEVDepth  \cite{li2023bevdepth}  & AOE$\downarrow$           & 15.3 &	18.2 &	\emph{12.5} &	\emph{52.5} &	130.2 &	83.1 &	\emph{69.0} &	\emph{82.8} &	- &	20.2 & \emph{48.0}  \\ 
        CRN-R50         \cite{kim2023crn}          & AOE$\downarrow$           & \emph{14.6} & \emph{15.3} & 13.0 & 60.0 & \emph{125.1} & \emph{68.3} & 82.0 & 114.0 & - & \emph{15.3}   & 54.2  \\ 
        \rowcolor[gray]{0.9}
        PAN-R50 (ours) & AOE$\downarrow$ & \textbf{9.7} & \textbf{9.9} & \textbf{8.3} & \textbf{50.2} & \textbf{123.8} & \textbf{59.6} & \textbf{44.7} & \textbf{46.4} & - & \textbf{10.7} & \textbf{40.4} \\
        \hline
        BEVDepth        \cite{li2023bevdepth}           & AAE$\downarrow$           & 21.5 &	\emph{20.1} &	30.5 &	\emph{8.7} &	\emph{35.5} &	25.3 &	\emph{22.2} &	\emph{1.0}	 & - &	- &	\emph{19.3}  \\ 
        CRN-R50       \cite{kim2023crn}            & AAE$\downarrow$           & \emph{19.7} & 20.7 & \textbf{17.8} & 10.0 & \textbf{35.2} & \emph{17.4} & \textbf{21.3} & 1.2 & - & -     & 19.7  \\ 
        \rowcolor[gray]{0.9}
        PAN-R50 (ours) & AAE$\downarrow$ & \textbf{18.7} & \textbf{19.5} & \emph{19.3} & \textbf{8.1} & 38.9 & \textbf{16.8} & 22.9 & \textbf{0.9} & - & - & \textbf{18.1} \\
        \hline
    \end{tabular}
\end{table*}

Additionally, we conducted a simplified study comparing two different distance ranges in Table \ref{tab:diff_dist}, the standard metrics are typically calculated for distances between 0 and 50 meters, but we separate this interval into two, 0 to 25 meters, and 25 to 50 meters, to better illustrate the robustness and viability of our results. Assuming a friction coefficient \begin{math}\mu\end{math} of 0.7 during braking on dry asphalt with ABS \cite{lyubenov2014study}, and based on the equation of motion, we can substitute these values into \eqref{motion}.
 \begin{equation}
\label{motion}
v_f^2 = v_0^2 + 2 \times (\mu \times g) \times d_b .
\end{equation}
Solving for distance, where \begin{math} v_f \end{math} is the final velocity, \begin{math} v_0 \end{math} is the initial velocity in \begin{math} (m/s) \end{math}, \begin{math} d_b \end{math} is the braking distance in \begin{math} (m) \end{math}, and \begin{math} g \end{math} is the gravitational acceleration multiplied by the friction coefficient in \begin{math} (m/s^2) \end{math}, we can estimate the braking distance of  \begin{math} 14.1 m \end{math} for a car moving at \begin{math} 50 km/h \end{math}, which is a reasonable estimation for the usual speed limit in the cities, considering only dry road conditions. It is also important to note that this distance should be significantly increased under adverse weather conditions, such as rain, potentially reaching up to twice this distance.

Since our study involves a comparison between human and robotic response times, we adopt an average human reaction time, \begin{math}t_r=1\end{math} second, as reported by Edquist et al. \cite{edquist2012effects}. However, this reaction time can be significantly reduced when applied to self-driving vehicles, where perception systems operate at much higher frame rates. For instance, in our work, the perception module operates at approximately 30 frames per second, enabling faster response times; we can use \eqref{time} to compute the estimated stopping or avoidance distance that corresponds to this reaction delay. This calculation provides a basis for comparing human-driven and autonomous vehicle response capabilities in dynamic environments.
 \begin{equation}
\label{time}
d_r = v_o  \times t_r .
\end{equation}

In this calculation, \begin{math} v_0 \end{math} is the initial velocity in \begin{math} (m/s) \end{math}, \begin{math} d_r \end{math} is the distance of reaction time in \begin{math} (m) \end{math}, and \begin{math}t_r\end{math} is the reaction time in \begin{math} (s) \end{math}. Using \eqref{time}, the estimated reaction distance a vehicle travels before it starts to respond is around 14 meters. When considering both stop distances, this results in an estimated total distance  \begin{math}d_t\end{math} of approximately 25 meters, using the sum of \begin{math}d_b\end{math} and \begin{math}d_r\end{math} components. To provide a meaningful comparison between our work and the main presented base models, we present results based on this 25-meter variation in Table \ref{tab:diff_dist}. 

In this experiment, we observed that by incorporating a limited velocity constraint and a sensor fusion approach, we achieved the best results within a range of 0 to 25 meters overall. However, our experiments also indicate a decline in performance metrics when considering greater distances (beyond 25 meters, up to 50 meters) and a significant gap between the desired perfect perception, at least for distances up to 25 meters. We are using the human reaction time because, as there are currently no fully automated self-driving cars, an autonomous system is expected to provide the reaction time needed for a human to take control in the event of a failure \cite{dixit2016autonomous}.

\begin{table*}[!htbp]
    \centering
    \renewcommand{\arraystretch}{1.2}
    \caption{Comparison of different distance intervals in validation, rain, and night sets to evaluate the safety aspect. We are considering two intervals due to the required braking time on a dry asphalt road applied at 50 km/h and around 1 second of reaction time. For all methods, the ResNet-50 version and the same image size were used.}
    \label{tab:diff_dist}
    \setlength{\tabcolsep}{5.2pt}
    \begin{tabular}{l|l|cc|cc|cc|cc|cc|cc}
        \hline
        Method & Version & \multicolumn{6}{c|}{Up to 25m} & \multicolumn{6}{c}{Greater than 25m (up to 50m)} \\
        \cline{3-14}
        & & \multicolumn{2}{c|}{Validation} & \multicolumn{2}{c|}{Rainy} & \multicolumn{2}{c|}{Night} & \multicolumn{2}{c|}{Validation} & \multicolumn{2}{c|}{Rainy} & \multicolumn{2}{c}{Night} \\
        & & NDS$\uparrow$ & mAP$\uparrow$ & NDS$\uparrow$ & mAP$\uparrow$ & NDS$\uparrow$ & mAP$\uparrow$ & NDS$\uparrow$ & mAP$\uparrow$ & NDS$\uparrow$ & mAP$\uparrow$ & NDS$\uparrow$ & mAP$\uparrow$ \\
        \hline
        BEVDepth \cite{li2023bevdepth} & R50 $(256 \times 704)$ & 54.8 & 49.0 & 57.2 & 51.1 & 28.8 & 23.8 & 33.0 & 16.2 & 34.8 & 18.1 & 17.4 & 8.5 \\
        CRN \cite{kim2023crn} & R50 $(256 \times 704)$ & 62.4 & 57.6 & 63.8 & 58.8 & \textbf{39.7} & \textbf{34.7} & 44.9 & 29.2 & 45.9 & 31.3 & 25.7 & 16.0 \\
        \rowcolor[gray]{0.9}
        PAN (ours) & R50 $(256 \times 704)$ & \textbf{64.3} & \textbf{58.3} & \textbf{66.5} & \textbf{60.3} & 38.8 & 32.3 &  \textbf{47.2} & \textbf{30.1} & \textbf{49.5} & \textbf{32.3} & \textbf{30.0} & \textbf{17.4} \\
        \hline
    \end{tabular}
\end{table*}

Additionally, although our perception system could operate at around 30 frames per second (FPS) for the R50 version, it is crucial to take into account additional factors that influence the self-driving car's reaction time, even considering the implementation in a fully autonomous system, such as the vehicle’s control system, actuators, communication delays, and other components. These elements collectively impact the overall response time of a self-driving car, making a direct comparison with human reaction times more complex, and to calculate a more complete estimation, it is necessary to evaluate these other aspects that supposedly should be shorter than the human reaction time.

Finally, to validate these results, we integrated a ResNet-50 backbone into our model and compared its performance against the CRN and BEVDepth baselines using the same image size of \begin{math} 256 \times 704\end{math} for a fair comparison. Additionally, in Table \ref{tab:diff_dist}, we included the BEVDepth model in the range-variation evaluation with different weather conditions because it is specifically designed to rely solely on camera input and follows a similar architectural approach to our PAN in the camera branch. By analyzing the metrics comparison using the camera only, 57.2/28.8 NDS, and our camera-radar approach, 66.5/38.8 NDS, for rain and night scenes, respectively, we can better assess the impact of incorporating radar data and sensor fusion in improving robustness in distinct scenarios, particularly under challenging conditions such as nighttime and rainy environments.

\subsection{ABLATION STUDY}

To evaluate the impact of different components in our architecture, we conducted a series of experiments by varying the size of the feature space. In large language models and other transformer-based approaches, this value is typically set to \begin{math} 128\end{math} \cite{de2022attention}. To analyze the effect of feature size on performance, we tested two additional values while maintaining a factor of 2 scaling up and down from this baseline. The results of these experiments are presented in Table \ref{tab:features}. Our findings indicate that, although altering the feature size does not result in significant performance, the best results are obtained using the default configuration of 
\begin{math}128\end{math} features, suggesting that this choice strikes an optimal balance between accuracy and efficiency, as this feature size increment changes only a small part of the architecture and consequently has a negligible impact on the inference time.

\begin{table}[h]
    \centering
    \renewcommand{\arraystretch}{1.1}
    \caption{Comparison using the ResNet-18 version, 32 epochs, and reduced feature size. The selected range represents a variation using the main value of 128, which is widely used in transformer architectures.}
    \label{tab:features}
    \setlength{\tabcolsep}{5pt}
    \begin{tabular}{l|cc|cc|cc}
        \hline
        Variation & \multicolumn{2}{c|}{Validation} & \multicolumn{2}{c|}{Rainy} & \multicolumn{2}{c}{Night} \\
        & NDS$\uparrow$ & mAP$\uparrow$ & NDS$\uparrow$ & mAP$\uparrow$ & NDS$\uparrow$ & mAP$\uparrow$ \\
        \hline
        PAN (\begin{math} f=192 \end{math}) & 54.0 & 43.4 & 55.3 & 45.5  & 33.1 & 24.7 \\
        PAN (\begin{math} f=128 \end{math}) & 54.1 & 43.7 & 55.3 & 45.7 & 33.2 & 25.0 \\
        PAN (\begin{math} f=64 \end{math}) & 54.0 & 43.5 & 55.4 & 45.6 & 33.3 & 25.2 \\
        \hline
    \end{tabular}
\end{table}

The second variation involved removing the convolutional layers in the last part of our feature enhancement block, keeping only the encoder, decoder, and self-attention branch (Fig. \ref{fig_att_block}) as the standard architecture for point-based 3D object detection could be reduced by reducing the convolutional layers. As the radars have a low resolution of points when compared to lidar sensors, which was the main focus of most of the works in the literature, we tried to remove these components in our work as well.  The results of this experiment are shown in Table \ref{tab:conv}

\begin{table}[h]
    \centering
    \renewcommand{\arraystretch}{1.1}
    \caption{Comparison using the ResNet-18 version, 32 epochs, and two different versions, with the convolutional layers and removing these layers.}
    \label{tab:conv}
    \setlength{\tabcolsep}{5pt}
    \begin{tabular}{l|cc|cc|cc}
        \hline
        Method & \multicolumn{2}{c|}{Validation} & \multicolumn{2}{c|}{Rainy} & \multicolumn{2}{c}{Night} \\
        & NDS$\uparrow$ & mAP$\uparrow$ & NDS$\uparrow$ & mAP$\uparrow$ & NDS$\uparrow$ & mAP$\uparrow$ \\
        \hline
        PAN (no conv) & 53.7 & 43.1 & 55.2 & 45.2  & 32.4 & 23.7 \\
        PAN (conv) & 55.3 & 44.4 & 57.6 & 47.6 & 33.7 & 23.8 \\
        \hline
    \end{tabular}
\end{table}

The result was worse than using a version with it, which means that to better capture the near correlations between objects, it is preferable to use the convolutional approach together with the self-attention branch. This is due to the fact that the transformers applied to computational vision are only comparable to convolutional layers using a large amount of data \cite{de2022attention}, and the nuScenes dataset, even being one of the biggest datasets for self-driving cars, has a small amount compared with the image classification databases, especially under adverse weather conditions.

\subsection{QUALITATIVE RESULTS}

\begin{figure}[!htbp]
  \centering
  \begin{minipage}[b]{0.44\linewidth}
    \centering
    \includegraphics[width=\linewidth]{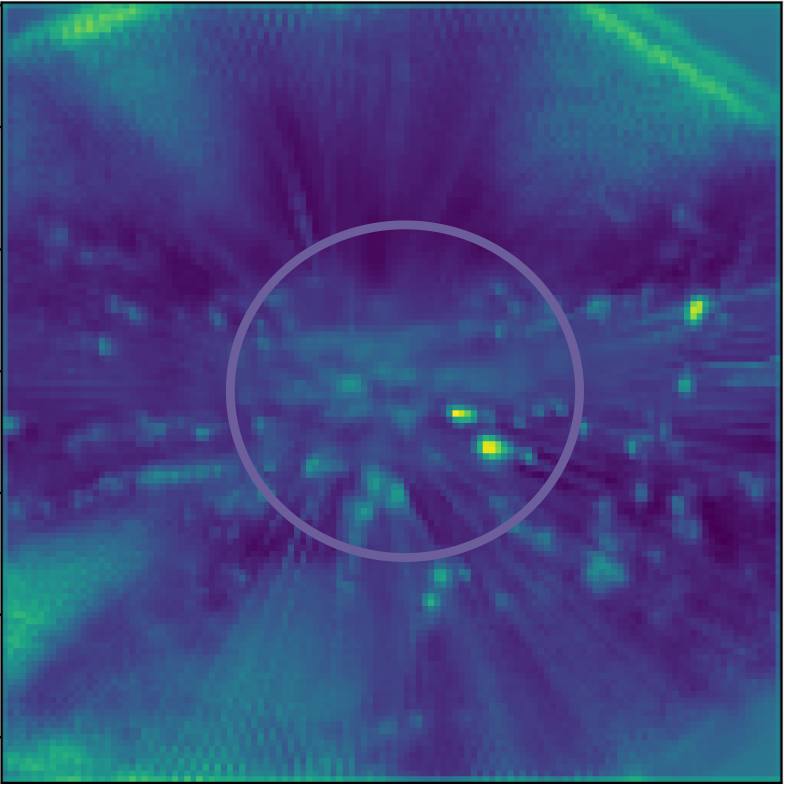}
    \\[-0.1em]
    {\small (a) CRN example 1}
  \end{minipage}
  \hfill
  \begin{minipage}[b]{0.44\linewidth}
    \centering
    \includegraphics[width=\linewidth]{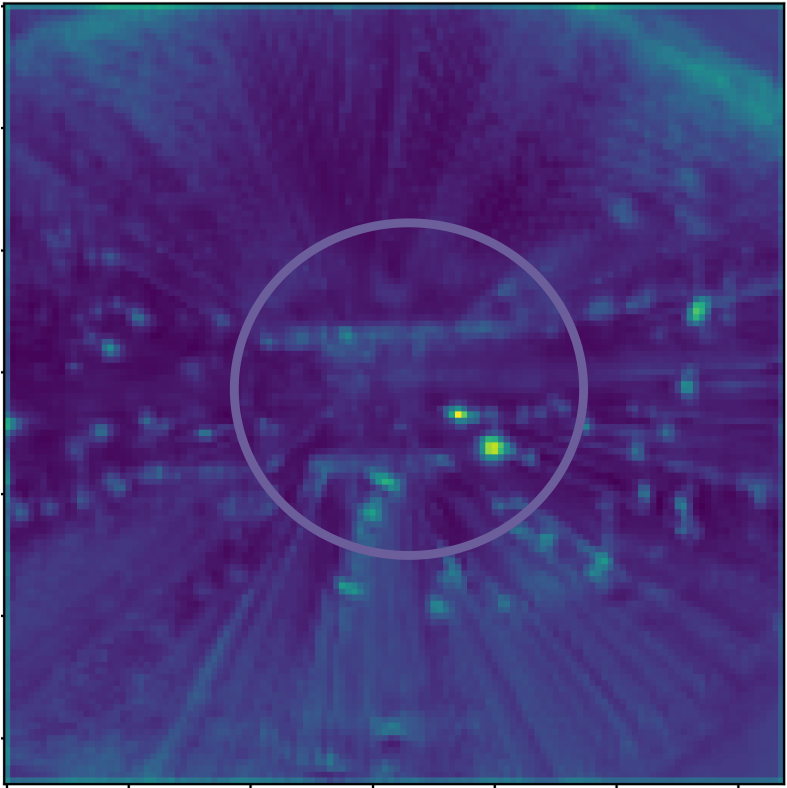}
    \\[-0.1em]
    {\small (b) PAN example 1}
  \end{minipage}

  \vspace{1em}

  \begin{minipage}[b]{0.44\linewidth}
    \centering
    \includegraphics[width=\linewidth]{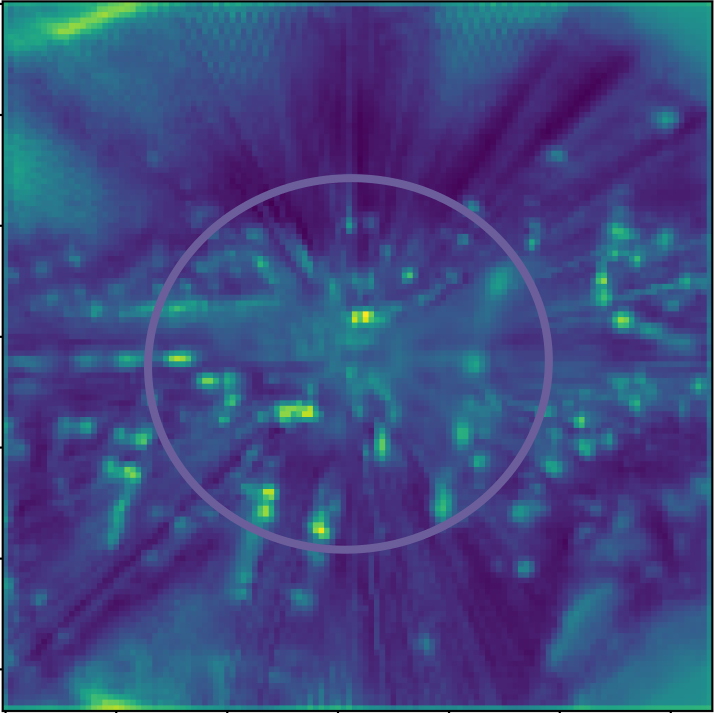}
    \\[-0.1em]
    {\small (c) CRN example 2}
  \end{minipage}
  \hfill
  \begin{minipage}[b]{0.44\linewidth}
    \centering
    \includegraphics[width=\linewidth]{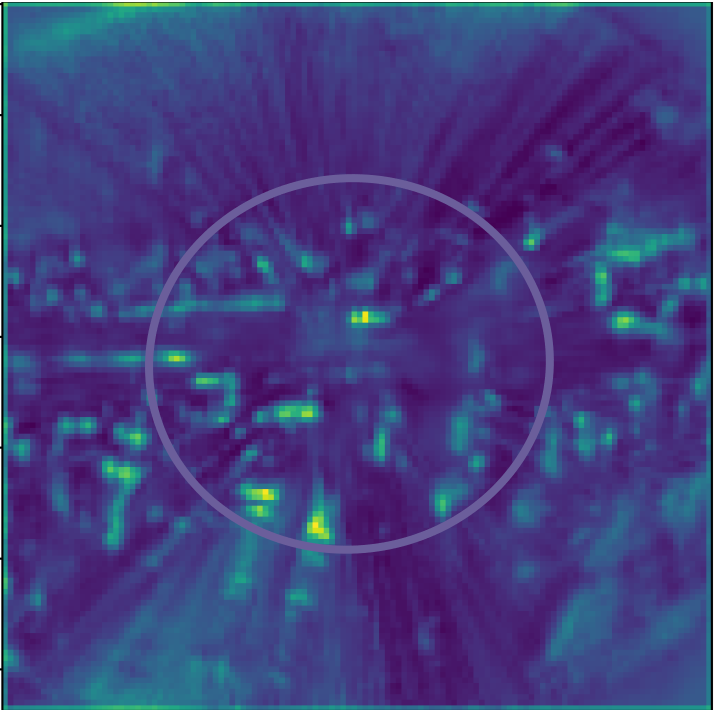}
    \\[-0.1em]
    {\small (d) PAN example 2}
  \end{minipage}

  \caption{Visualization for the feature space. The feature maps shown in (a) and (b) and (c) and (d) are the models under two different scenes, where the lighter areas represent the regions most used for detection.}
  \label{features}
\end{figure}

To provide a clearer visualization of the differences between our proposed model and the primary baseline, CRN, we conducted a first comparative qualitative analysis by plotting the bird’s-eye-view feature space using both architectures to observe how our model is improving the feature space. We aggregated all feature values in the BEV space to carry out this experiment. Specifically, we utilized a total of 128 features, with a spatial grid size of \begin{math} 128 \times 128 \end{math} to cover the distance of 50 meters in each axis. By summing all the feature values for each sample within this space, we aimed to gain insights into how the learned representations differ between the two approaches. The resulting visualization is presented in Fig. \ref{features}, which illustrates the improvements achieved by our model in terms of feature space.

In Fig. \ref{features}, it is possible to observe the difference between our approach and the base model, where our approach can focus more on the objects of interest, reducing the weights assigned to roads and non-desirable objects. The feature analysis can be exemplified using Fig. \ref{features} (a) and Fig. \ref{features} (b) where the corners are less highlighted than the cars and pedestrians and through Fig. \ref{features} (c) and Fig. \ref{features} (d) where it is possible to observe a clearer distinction between objects of interest and the background. For this visual experiment, we use a sample-based normalization where higher values in each sample are depicted as lighter regions in the image.

\begin{figure*}[!t]
  \centering

  \begin{minipage}[b]{0.75\textwidth}
    \centering
    \includegraphics[width=\linewidth]{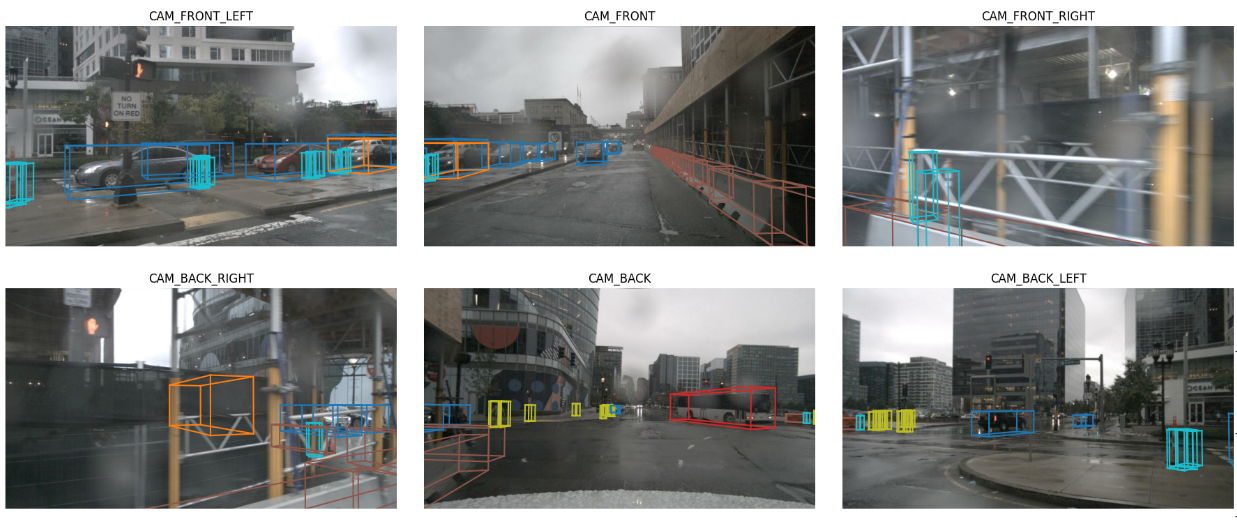}
    \\[-0.5em]
    {\small (a) CRN camera views}
  \end{minipage}

  \vspace{1em}

  \begin{minipage}[b]{0.75\textwidth}
    \centering
    \includegraphics[width=\linewidth]{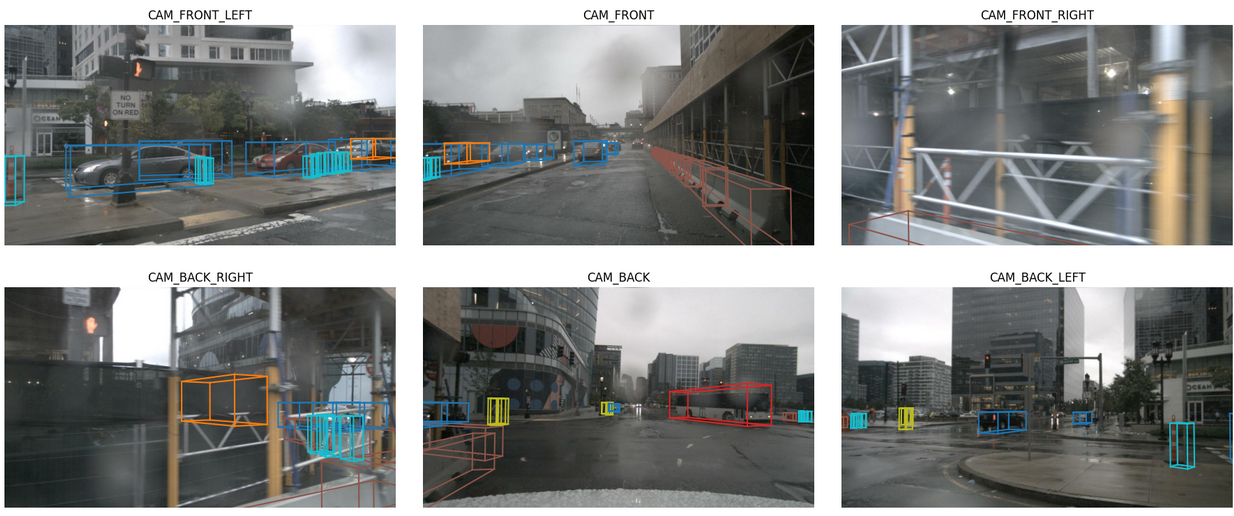}
    \\[-0.5em]
    {\small (b) PAN camera views}
  \end{minipage}

  \vspace{1em}

  \begin{minipage}[b]{0.75\textwidth}
    \centering
    \includegraphics[width=\linewidth]{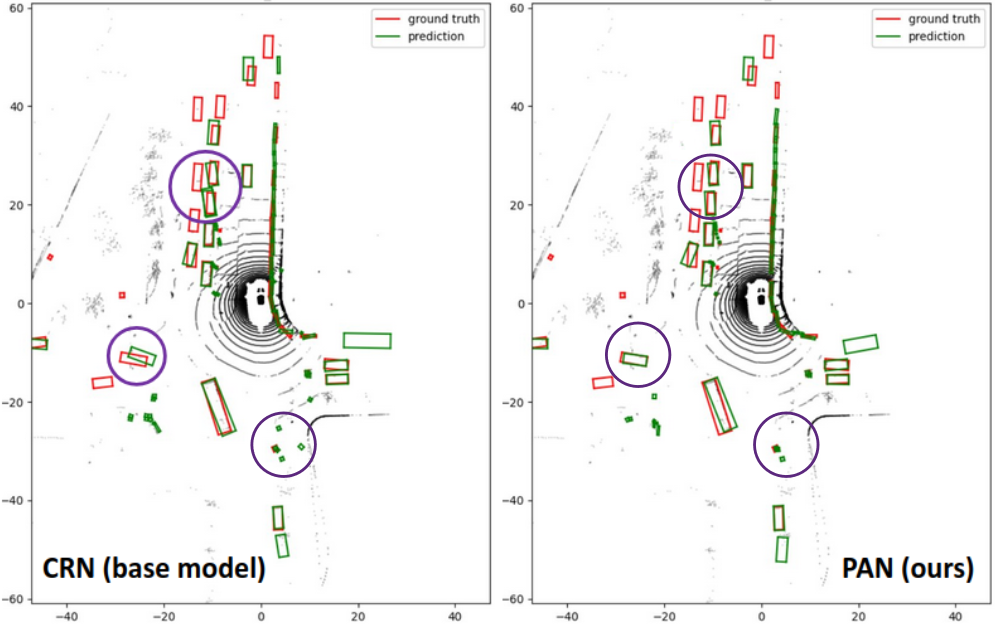}
    \\[-0.5em]
    {\small (c) BEV view}
  \end{minipage}

  \caption{Visualization for the qualitative results. The bounding boxes shown in (a) and (b) are the CRN and PAN predictions, respectively, where yellow indicates pedestrians, red buses, orange trucks, cyan traffic cones, salmon barriers, and blue cars. Panel~(c) shows the BEV perspective for CRN (base model) and PAN (ours), with the main improvements highlighted in purple. It is also possible to observe that PAN can correct wrong class attributions.}
  \label{fig:qual}
\end{figure*}

In our second qualitative experiment, to provide a comprehensive insight into the performance of our algorithm in the real world, we provided detailed visualizations showcasing the influence of our architecture on the 3D object detection problem and the main PAN contributions. For a comparison with more insights, the Fig. \ref{fig:qual} visualization also includes an image generated using our main baseline CRN. A rainy scene was selected for better visualization of the performance under adverse weather conditions.

Fig. \ref{fig:qual} depicts the CRN and PAN detections using all camera views (respectively Fig. \ref{fig:qual}(a) and Fig. \ref{fig:qual}(b)), where all the bounding boxes shown are the predictions using the ResNet-50 version of each 3D object detection algorithm. Furthermore, Fig. \ref{fig:qual}(c) presents a BEV visualization of CRN and PAN models, utilizing the available ground truth lidar points as background because of their density of points and better visualization of the object's shape. The BEV scene of Fig. \ref{fig:qual}(c) demonstrates that our model, PAN, outperforms the baseline model under challenging visibility conditions, as demonstrated by the more precise 3D bounding boxes it generates. The purple circles indicated in the Fig. \ref{fig:qual} show examples where PAN fits the ground truth more accurately than CRN. Additionally, it can be observed that our model will increase the metrics related to the orientation, whereas the baseline model occasionally produces erroneous orientation predictions. These results validate the effectiveness of our attention-based approach, which also reduces false positives.

\section{CONCLUSION}\label{sec:conclusion}
In this paper, we propose PAN, a new state-of-the-art architecture for 3D object detection using a camera-radar approach under the nuScenes dataset. We tested the results using a ResNet-50 and ResNet-18 image backbone for a fair comparison with older models, and we provided a study using the two main bird's-eye view baselines, CRN and BEVDepth models, under regular and adverse weather conditions of nighttime and rainy scenes.

To achieve a good balance between metric performance and inference time, we used a pillars-based self-attention approach in our main enhancement block, which allowed us to more effectively utilize the global information and the radar features, leading to a faster and more accurate approach than the baseline models presented in the literature so far. The observed improvements were more significant for features related to the orientation and attribute characteristics of the 3D detected objects.

Overall, our model can outperform the baseline in our best-presented version, the ResNet-50 version, at 43\% for inference time while reaching the 58.2 scores for the NDS metric. Furthermore, this paper provides a study on the safety of using a camera-radar sensor fusion approach to autonomous vehicles and provides an estimate of some desirable factors, such as a minimum distance for perfect detection and performance reduction when the vehicle is under adverse weather conditions.

Even though our work is currently a state-of-the-art method, there is a long way to go to improve the metrics for real-world applications. We observe that, at least for 25-meter detection, it is desirable for the model to be able to make human-level detections to avoid accidents, following the human-annotated information in the nuScenes dataset that provided our ground truth. We also realize that, beyond other approaches, the radar used in the nuScenes dataset is an old model with low resolution. Currently, there are better radar models that can also be tested to reduce this difference.

\bibliography{src/ref}
\begin{IEEEbiography}[{\includegraphics[width=1in,height=1.3in,clip,keepaspectratio]{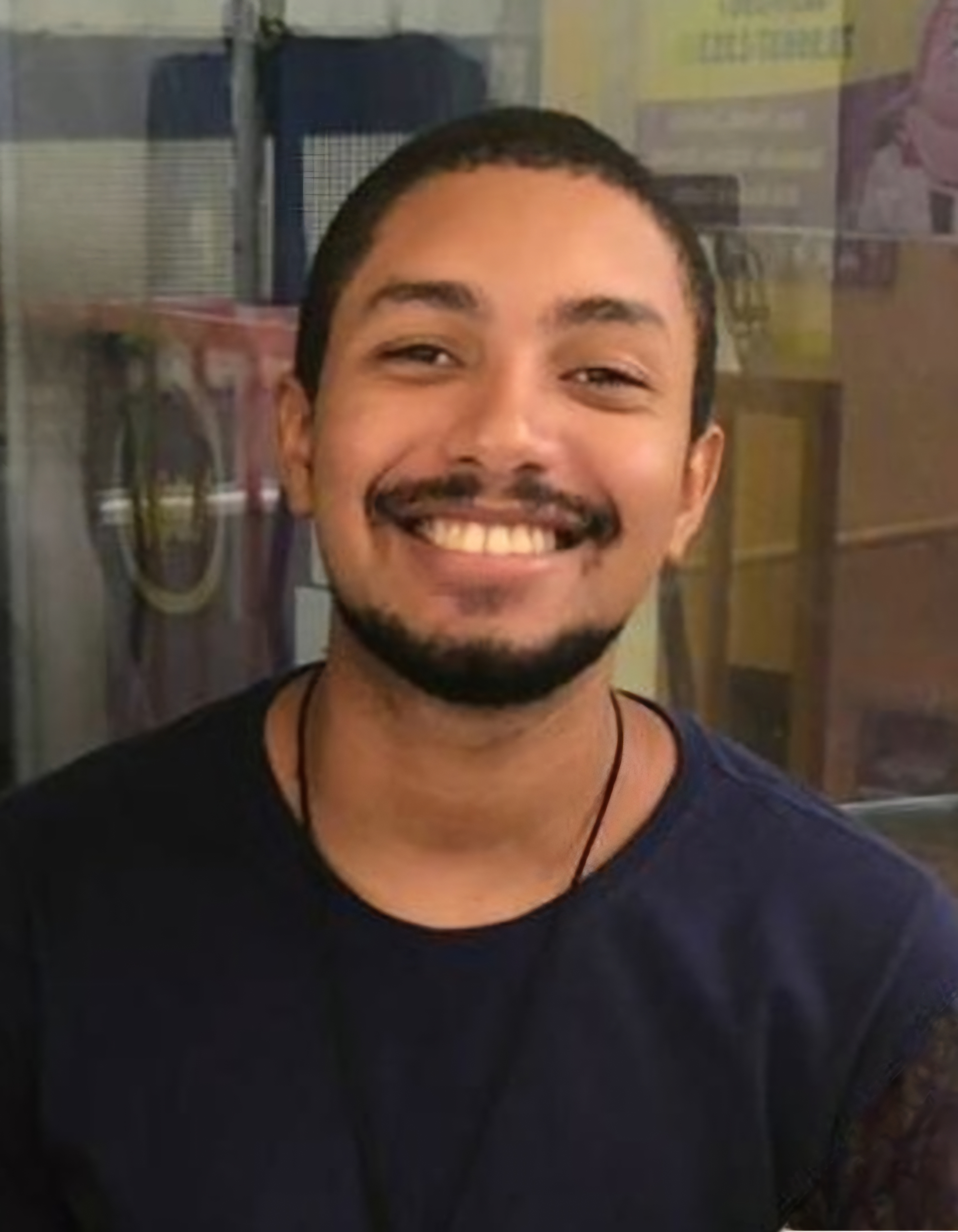}}]{Ruan Bispo } 
received the BA degree in electronic engineering from the Federal University of Sergipe in 2020, the MSE degree in dynamic systems from the University of S\~ao Paulo in 2023, and is currently a first-year PhD student in computer vision and artificial intelligence at the University of Limerick. He was a researcher at Ford Motor Company (2 years) and a data scientist at Samsung (2 years). His research interests include robotics, autonomous vehicles, control, and AI.
\end{IEEEbiography}

\begin{IEEEbiography}[{\includegraphics[width=1in,height=1.3in,clip,keepaspectratio]{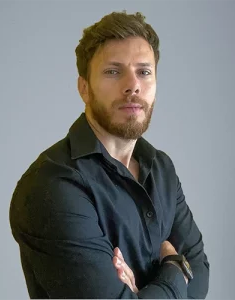}}]{Dane Mitrev } 
is a Senior ML Engineer at Provizio, specializing in the processing of point cloud data from radar and lidar sensors, and the application of machine learning techniques to vehicle sensor data. He is working in the intersection of Machine Learning and Computer Vision for perception and sensor fusion. With a wealth of experience in the field, he has tackled a wide range of topics including point cloud denoising, super-resolution, point cloud segmentation, scene flow, accumulation, object detection, and tracking.
\end{IEEEbiography}

\begin{IEEEbiography}[{\includegraphics[width=1in,height=1.3in,clip,keepaspectratio]{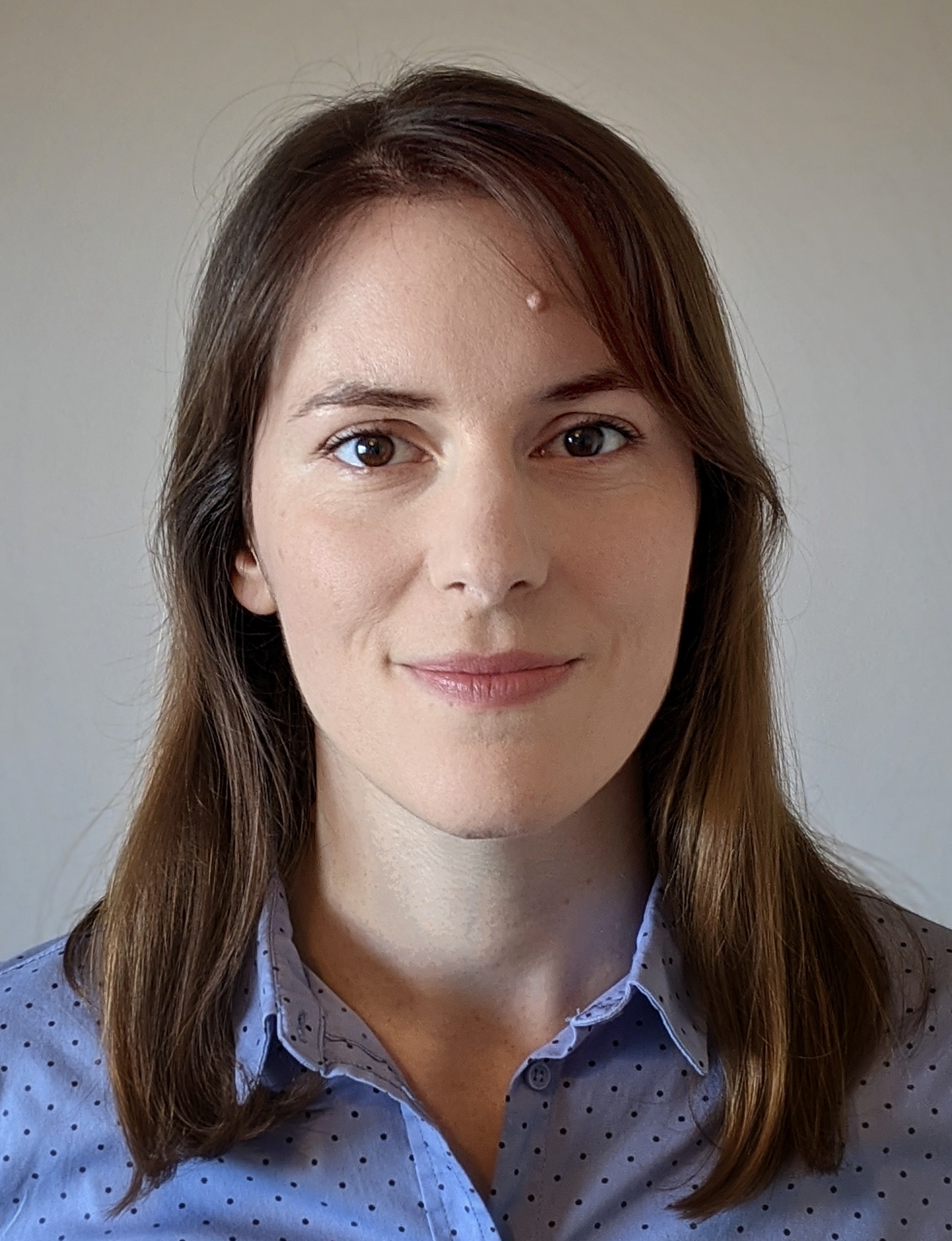}}]{Letizia Mariotti } 
received the BSc degree in Physics from the University of Trieste in 2010, the MSc degree in Astrophysics and Cosmology from the University of Trieste in 2013, and the PhD degree in Physics from the National University of Ireland, Galway in 2017.
From 2017 she worked as a Computer Vision Research Engineer in Valeo Vision Systems, Ireland. She joined Provizio as a Senior Computer Vision Engineer in 2021.
\end{IEEEbiography}

\begin{IEEEbiography}[{\includegraphics[width=1in,height=1.3in,clip,keepaspectratio]{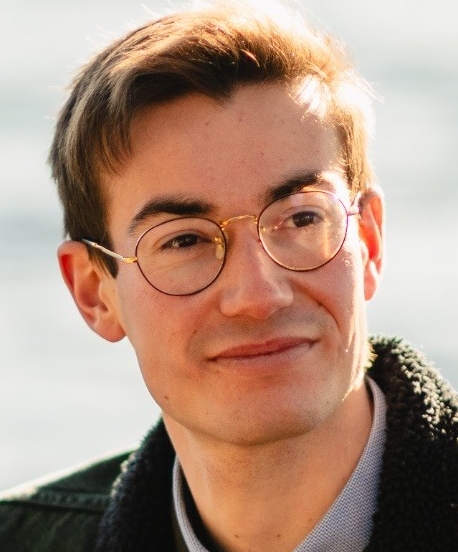}}]{Clément Botty } 
is a Machine Learning engineer focused on radar perception in the edge. With a background in the French Armed Forces Ministry, he has led both engineering initiatives and product development for applications serving over 20,000 users. His expertise spans radar perception services, machine learning model optimization and product management. Outside of work, Clément is passionate about tech, cycling, and surfing the coasts of Brittany—occasionally swapping the board for a kite.
\end{IEEEbiography}

\begin{IEEEbiography}[{\includegraphics[width=1in,height=1.3in,clip,keepaspectratio]{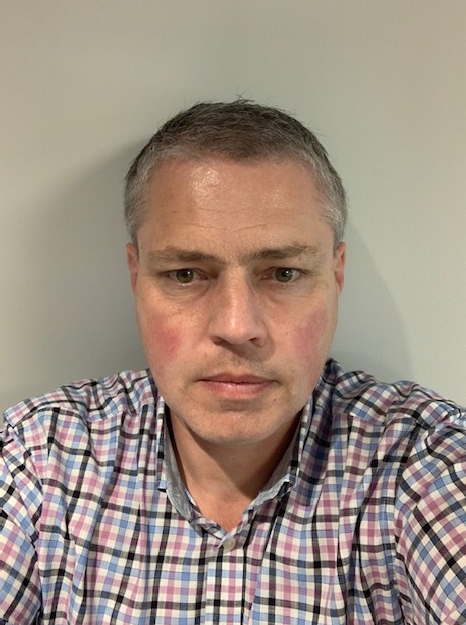}}]{Denver Humphrey } 
received the BEng degree in Electrical and Electronic Engineering and the PhD degree in High Frequency Electronics from the Queen's University of Belfast in 1993 and 1996 respectively. He has previously worked as a Principal design engineer at Merlin Microwave, Celeritek, TDK and Arralis where he worked on various projects for the space sector. He joined Provizio as a Principal design engineer in 2019.
\end{IEEEbiography}

\begin{IEEEbiography}[{\includegraphics[width=1in,height=1.3in,clip,keepaspectratio]{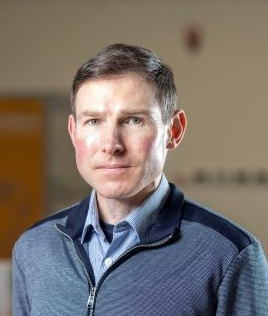}}]{\textbf{Anthony Scanlan }} received the B.Sc. degree in experimental physics from the National University of Ireland Galway, Galway, Ireland, in 1998 and the M.Eng. and Ph.D. degrees in electronic engineering from the University of Limerick, Limerick, Ireland, in 2001 and 2005, respectively. He is currently a Senior Research Fellow at the Dept. of Electronic \& Computer Engineering, University of Limerick, Ireland, and has been the principal investigator on several research projects in the areas of signal processing and data converter design. His current research interests are in the areas of artificial intelligence, computer vision, and industrial and environmental applications. 
\end{IEEEbiography}

\begin{IEEEbiography}[{\includegraphics[width=1in,height=1.3in,clip,keepaspectratio]{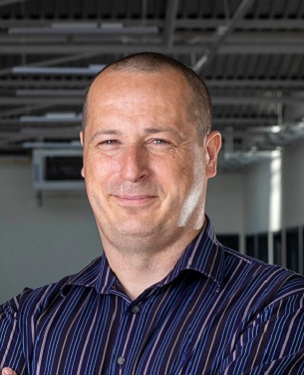}}]{Ciarán Eising } received the BE in Electronic and Computer Engineering and a PhD from the NUI Galway, in 2003 and 2010, respectively. From 2009 to 2020, he was a Computer Vision Architect \& Senior Expert with Valeo. In 2020, he joined the University of Limerick as an Associate Professor.
\end{IEEEbiography}
\end{document}